\documentclass{article}

\usepackage{microtype}
\usepackage{graphicx}
\usepackage{subfigure}
\usepackage{booktabs} 

\usepackage[accepted]{icml2024}

\usepackage{amsmath}
\usepackage{amssymb}
\usepackage{mathtools}
\usepackage{amsthm}

\usepackage{url}
\usepackage{graphicx}
\usepackage{amsmath}
\usepackage{amssymb}
\usepackage{booktabs}

\usepackage{xcolor} 
\usepackage{times}
\usepackage{epsfig}
\usepackage{graphicx}
\usepackage{amsmath}
\usepackage{amssymb}
\usepackage[bb=dsserif]{mathalpha}
\usepackage{bm}

\usepackage{booktabs,colortbl,tabularx}
\usepackage{pifont}%

\usepackage{mathtools}
\usepackage{commath}
\usepackage{algorithm}

\usepackage{multirow}
\usepackage{comment} 
\usepackage{enumitem}

\definecolor{citecolor}{HTML}{2980b9}
\definecolor{linkcolor}{HTML}{c0392b}

\usepackage[pagebackref=true,breaklinks=true,colorlinks,bookmarks=false,citecolor=citecolor,linkcolor=linkcolor]{hyperref}

\definecolor{Gray}{gray}{0.9}

\renewcommand{\paragraph}[1]{\noindent {\bf #1}}

\newcommand{\cmark}{\ding{51}}%
\newcommand{\xmark}{\ding{55}}%

\usepackage[capitalize,noabbrev]{cleveref}

\theoremstyle{plain}

\theoremstyle{definition}

\theoremstyle{remark}

\icmltitlerunning{DMT-JEPA: Discriminative Masked Targets for Joint-Embedding Predictive Architecture}

\begin{document}

\twocolumn[
\icmltitle{DMT-JEPA: Discriminative Masked Targets for Joint-Embedding \\ Predictive Architecture}




\begin{icmlauthorlist}
\icmlauthor{Shentong Mo}{cmu,mbz}
\icmlauthor{Sukmin Yun}{hyu}
\end{icmlauthorlist}

\icmlaffiliation{cmu}{Carnegie Mellon University}
\icmlaffiliation{mbz}{Mohamed bin Zayed University of Artificial Intelligence}
\icmlaffiliation{hyu}{Hanyang University ERICA}

\icmlcorrespondingauthor{Sukmin Yun}{sukminyun@hanyang.ac.kr}

\icmlkeywords{Machine Learning, ICML}

\vskip 0.3in
]



\printAffiliationsAndNotice{}  

\begin{abstract}

The joint-embedding predictive architecture (JEPA) recently has shown impressive results in extracting visual representations from unlabeled imagery under a masking strategy.
However, we reveal its disadvantages, notably its insufficient understanding of local semantics.
This deficiency originates from masked modeling in the embedding space, resulting in a reduction of discriminative power and can even lead to the neglect of critical local semantics.
To bridge this gap, we introduce DMT-JEPA, a novel masked modeling objective rooted in JEPA,
specifically designed to generate discriminative latent targets from neighboring information.
Our key idea is simple: we consider a set of semantically similar neighboring patches as a target of a masked patch.
To be specific, the proposed DMT-JEPA (a) computes feature similarities between each masked patch and its corresponding neighboring patches to select patches having semantically meaningful relations, and (b) employs lightweight cross-attention heads to aggregate features of neighboring patches as the masked targets.
Consequently, DMT-JEPA demonstrates strong discriminative power, offering benefits across a diverse spectrum of downstream tasks.
Through extensive experiments, we demonstrate our effectiveness across various visual benchmarks, including ImageNet-1K image classification, ADE20K semantic segmentation, and COCO object detection tasks.
Code is available at: \url{https://github.com/DMTJEPA/DMTJEPA}.

\end{abstract}

\vspace{-1.5em}
\section{Introduction}

The success of self-supervised learning (SSL) frameworks~\citep{chen2020simple,chen2021simsiam,he2019moco,grill2020bootstrap}, especially in harnessing vast reservoirs of unlabeled images, has been undeniable in the computer vision community.
Model architectures like the Vision Transformer (ViT;~\cite{dosovitskiy2021an}) have consistently garnered significant attention, and initial attempts at seamless integration with SSL have indeed demonstrated potential~\citep{chen2021mocov3,xie2021self-supervised,caron2021emerging}.
In particular, Masked autoencoder (MAE;~\cite{he2021masked}), which reconstructs missing patches on pixel space, has achieved advanced success in various visual downstream tasks, such as image classification, object detection, and semantic segmentation.

Recently, the image-based joint-embedding predictive architecture (I-JEPA;~\citet{assran2023self}) has shown promising results in learning self-supervised representations by leveraging a masking strategy to reconstruct representations of masked patches.
Specifically, I-JEPA uses a masked image to predict the representations of various unmasked blocks located in the same image rather than focusing on the unmasked pixels, thereby demonstrating superior efficacy and efficiency.
Nevertheless, we observed that this approach often results in performances that are less compatible during fine-tuning, especially when compared to the pixel reconstruction method ({\it e.g.,} MAE), and also results in indistinct attention maps, as illustrated in Figure~\ref{fig: exp_vis_attention}.
These indistinctnesses can be attributed to insufficient discriminative power, which is crucial for a deep understanding of local semantics.

One challenge posed by existing SSL approaches built upon ViTs is the potential lack of local semantics in the extracted representations from disjoint input patches, where local semantics are naturally intertwined within the image patches.
This inspiration leads us to integrate explicit processing of local semantics, aiming to produce discriminative latent targets that enhance masked modeling for predicting the embedding of masked images from unmasked ones.
To avoid the creation of non-discriminate latent targets, our principal strategy involves generating a semantically meaningful target for each masked patch. 
This strategy, inspired by the prior work of~\cite{yun2022patch}, utilizes the similarities among patches within a specific neighborhood to capture local semantics comprehensively.
During pre-training, we leverage these self-supervised latent targets to capture local semantics, thereby offering benefits for various tasks, including dense prediction and image classification.

In this paper, we introduce the Discriminative Masked Targets for Joint-Embedding Predictive Architecture (DMT-JEPA), a novel self-supervised representation learning framework focused on latent reconstruction for a masked image.
Our goal is to generate discriminative latent targets that can capture local semantics that can serve as an alternative masked modeling objective, which can be incorporated with the joint-embedding predictive architecture \citep{lecun2022path}.
To this end, we propose Masked Semantic Neighboring to find semantically similar neighboring patches for masked patches and Local Aggregation Target to generate the discriminative dense targets from them.
Specifically, Masked Semantic Neighboring computes feature similarities between each masked patch and its corresponding neighboring patches to select semantically similar patches, and Local Aggregation Target employs lightweight cross-attention heads to aggregate features of chosen neighboring patches as the latent targets for masked patches.
Consequently, the proposed DMT-JEPA not only exhibits superior efficiency comparable to I-JEPA~\cite{assran2023self}, but also achieves enhanced efficacy in learning self-supervised representations by capturing discriminative embeddings as illustrated in Figure~\ref{fig: exp_vis_attention}. 
This significant improvement is advantageous across a broad spectrum of tasks, encompassing image classification as well as dense prediction downstream tasks such as semantic segmentation and object detection.

To demonstrate the effectiveness of DMT-JEPA, we conduct extensive downstream experiments after pre-training ViT-B/16 and ViT-L/16~\citep{dosovitskiy2021an} on ImageNet-1k; these experiments include ImageNet-1K image classification, COCO object detection, ADE20K semantic segmentation, DAVIS video segmentation, and Clevr local prediction benchmarks.
Our experimental results demonstrate that the DMT-JEPA improves the performance of I-JEPA with a large margin and even outperforms other SSL baselines~\citep{he2021masked, chen2021mocov3, bao2021beit} on various benchmarks; 
for example, our method achieved +1.4 mIoU ({\it i.e.,} 47.6 → 49.0) on ADE20K semantic segmentation, +1.7 $(\mathcal{J} \& \mathcal{F})_m$ ({\it i.e.,} 56.6 → 58.3) on DAVIS video segmentation, +1.0 AP$^{\mathtt{box}}$ ({\it i.e.,} 49.9 → 50.9) on COCO object detection, and +1.1 AP$^{\mathtt{mask}}$ ({\it i.e.,} 44.5 → 45.6) on COCO instance segmentation.
Furthermore, we observed that DMT-JEPA can even outperform MAE in ImageNet fine-tuning experiments, achieving scores of 84.6 (on ViT-B/16) and 86.6 (on ViT-L/16), in contrast to MAE's scores of 83.6 (on ViT-B/16) and 85.9 (on ViT-L/16). This demonstrates that the proposed method not only benefits dense representation learning but also enhances the quality of global image representations.

Overall, our work underscores the effectiveness of utilizing discriminative targets for masked modeling within the embedding space. We aspire that our contributions will inspire researchers to further investigate latent reconstruction-based representation learning from a self-supervised perspective.

\section{Method}

\begin{figure*}[t]
\centering
\includegraphics[width=0.90\linewidth]{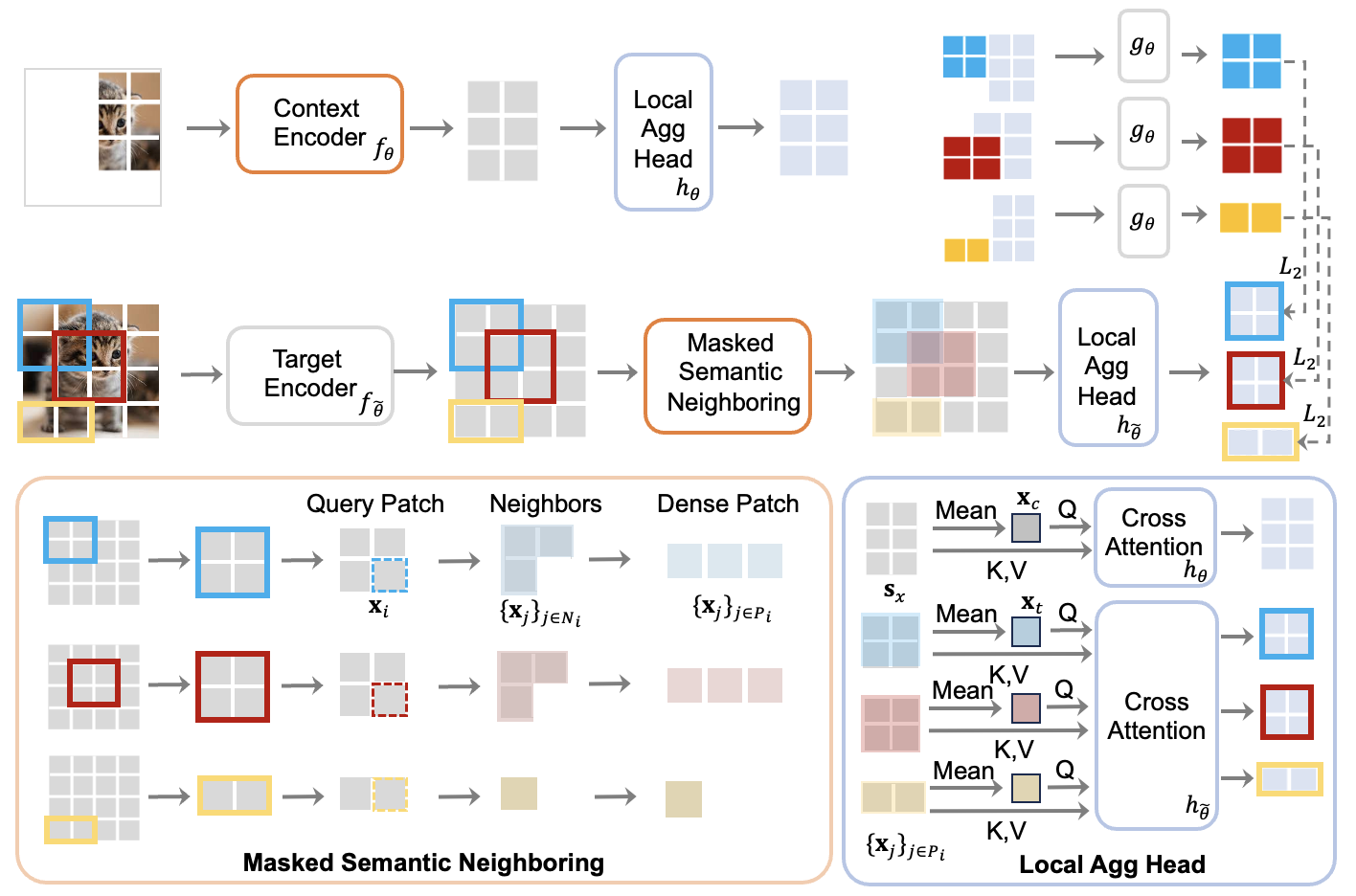}
\vspace{-1.0em}
\caption{{\bf Illustration of the proposed novel masked modeling framework (DMT-JEPA), rooted in I-JEPA, designed for 
discriminative latent targets from neighboring information.
}
The Masked Semantic Neighboring module computes the dense semantic similarity between the query patch and its neighboring patch based on representations from the target encoder $f_{\tilde{\theta}}(\cdot)$ to select semantically similar patches from the neighborhood.
Then Local Aggregation Target module composed of a context patch aggregation head $h_\theta(\cdot)$ and a target patch aggregation head $h_{\tilde{\theta}}(\cdot)$, aggregates target features of selected patches using cross-attention to construct dense targets.
Finally, the model is optimized by the average $L_2$ distance between
the predicted dense representations and the target dense representation.
}
\vspace{-1.0em}
\label{fig: main_img}
\end{figure*}

In this section, we present a novel masked modeling framework, coined DMT-JEPA, designed for the joint-embedding predictive architecture to enhance understanding local semantics within images.
Our key idea is that semantically similar representations can provide local semantics as a masked modeling objective by enforcing them to have similar representations. 
We first provide preliminaries in Section~\ref{sec:prelim} and then present details of two modules, Masked Semantic Neighboring in Section~\ref{sec:msn} and Local Aggregation Target in Section~\ref{sec:lat}.
Figure~\ref{fig: main_img} shows an overall illustration of the proposed method, DMT-JEPA.

\subsection{Preliminaries}\label{sec:prelim}

We first describe the problem setup and notations and then revisit the Image-based Joint-Embedding Predictive Architectures (I-JEPA; \citet{assran2023self}), which is a self-supervised visual representation learning under masked modeling.

\noindent\textbf{Problem Setup and Notations.}
Given an image with a dimension of $3\times H\times W$ and a patch resolution of $P$, our goal is to learn a masked autoencoder framework with an encoder $f_e(\cdot)$ and a decoder $f_d(\cdot)$ to recover the masked patches using unmasked ones. 
We formally denote patch embeddings of raw input via each linear projection layer, \textit{i.e.}, $\mathbf{x}\in\mathbb{R}^{N\times D}$,
$H$ and $W$ are the height and width of each image, and $D$ is the dimension of features.
Note that $N = H/P\times W/P$ and $N$ is the total number of patches.

\noindent\textbf{Masked Autoencoder.}
To address the masked image modeling problem, MAE~\citep{he2021masked} first applied a random masking set $M$ along the total number of patches, and then an encoder to extract features from unmasked patches.
Finally, unmasked embeddings and masked tokens were concatenated into a decoder to recover the raw pixels of masked patches.
The vanilla masking loss for each image is calculated with the mean square loss between the targeted $\mathbf{p}_i$ and predicted normalized pixels $\hat{\mathbf{p}}_i$ as:  
\begin{equation}
    \mathcal{L}_{\mathtt{MAE}} = \dfrac{1}{|M|}\sum_{i\in M} ||\mathbf{p}_i - \hat{\mathbf{p}}_i||_2^2,
\end{equation}
where $|M|$ denotes the total number of masked patches in the masking set $M$.

\noindent\textbf{Image-based Joint-Embedding Predictive Architecture.}
To tackle the masked image modeling task, I-JEPA~\citep{assran2023self} introduced a context encoder $f_\theta(\cdot)$, a target encoder $f_{\tilde{\theta}}(\cdot)$, and a predictor $g_\theta(\cdot)$, to predict the $M$ target block representations $\mathbf{s}_y(1),...,\mathbf{s}_y(M)$ given the output of the context encoder, $\mathbf{s}_x$.
For a target block $\mathbf{s}_{y_i}$ corresponding to a target mask $\mathcal{B}_i$, the predictor $g_\theta(\cdot, \cdot)$ takes as input the output of the context encoder $\mathbf{s}_x$ and a mask token for each patch to predict $\{\mathbf{m}_j\}_{j\in\mathcal{B}_i}$, and outputs the patch-level prediction $\{\hat{\mathbf{s}}_{y_j}\}_{j\in\mathcal{B}_i}$, that is, $\{\hat{\mathbf{s}}_{y_j}\}_{j\in\mathcal{B}_i}=g_\theta(\mathbf{s}_x,\{\mathbf{m}_j\}_{j\in\mathcal{B}_i})$.
The masking objective is optimized by the average $L_2$ distance between the predicted patch-level representations $\hat{\mathbf{s}}_{y_j}$ and the target patch-level representation $\mathbf{s}_{y_j}$, which is formulated as:
\begin{equation}
    \mathcal{L}_{\mathtt{I}\text{-}\mathtt{JEPA}} = \dfrac{1}{|M|}\sum_{i=1}^M \sum_{j\in \mathcal{B}_i} ||\mathbf{s}_{y_j} - \hat{\mathbf{s}}_{y_j}||_2^2,
\end{equation}
where $|M|$ denotes the total number of target blocks, and $\mathcal{B}_i$ is the mask corresponding to the $i$-th target block.

\subsection{DMT-JEPA: Masked Semantic Neighboring}\label{sec:msn}

However, a masked modeling target in the representation space like I-JEPA could pose a challenge in terms of missing local semantics if the target patch-level representations $\mathbf{s}_{y_j}$ were less discriminative among themselves. 
As shown in Figure~\ref{fig: exp_vis_attention}, we also observed that I-JEPA often generates indistinct attention maps, and it arguably indicates its deficiency in comprehending local semantics.
To tackle this, we aim to generate target representations capturing local semantics that can serve as an alternative masked modeling objective, which can be incorporated with the joint-embedding predictive architecture \citep{lecun2022path}.
We note that prior investigation on patch-level representation learning \citep{yun2022patch} inspires us to explore similarities among patch-level representations located in a neighborhood.
To this end, we propose Masked Semantic Neighboring module to find semantically similar neighboring patches for masked patches and Local Aggregation Target module (see Section~\ref{sec:lat}) to make them have similar target representations.

\noindent\textbf{Masked Semantic Neighboring.}
For patches in a given masked block, we aim to find their neighboring patches semantically similar, as neighboring patches often share a semantic context.
In order to sample semantically similar patches from the neighborhood $\mathcal{N}_i$, we compute the dense semantic similarity $d(i,j)$ between the query patch $\mathbf{x}_i$ and its neighboring patch $\mathbf{x}_j$ for all $j\in\mathcal{N}_i$ based on representations from the target encoder $f_{\tilde{\theta}}(\cdot)$, which is formulated as:
\begin{equation}
\begin{aligned}
    d(i,j) = \dfrac{f_{\tilde{\theta}}(\mathbf{x}_i)^\top f_{\tilde{\theta}}(\mathbf{x}_j)}{\|f_{\tilde{\theta}}(\mathbf{x}_i)\|_2\|f_{\tilde{\theta}}(\mathbf{x}_j)\|_2},
\end{aligned}
\end{equation}
where $f_{\tilde{\theta}}(\mathbf{x}_i),f_{\tilde{\theta}}(\mathbf{x}_j)\in\mathbb{R}^{1\times D}$, and $\|\cdot\|_2$ denotes the $\ell_2$-norm operator.
With the computed similarity scores, we apply a ranking on the neighboring patches $\{\mathbf{x}_j\}_{j\in\mathcal{N}_i}$ and select a set of dense patches $\{\mathbf{x}_j\}_{j\in\mathcal{P}_i}$ with top-$k$ highest similarities, where $\mathcal{P}_i$ denotes a set of dense patch indices in the neighborhood, and $k$ is the number of dense patches, \textit{i.e.}, $k=|\mathcal{P}_i|$.
Unless stated otherwise, we use $k = 4$ for our experiments.

\subsection{DMT-JEPA: Local Aggregation Target}
\label{sec:lat}

We remark that our goal is to generate target representations capturing local semantics and discriminative among themselves.
With the benefit of the selected neighboring patches having similar semantics,
we introduce Local Aggregation Target module composed of a context patch aggregation head $h_\theta(\cdot)$ and a target patch aggregation head $h_{\tilde{\theta}}(\cdot)$.
Specifically, we aggregate target representations of selected patches $\{f_{\tilde{\theta}}(\mathbf{x}_j)\}_{j\in\mathcal{P}_i}$ using cross-attention head $h_\theta(\cdot)$ to construct dense target $\mathbf{s}_i^{\mathtt{LAT}}$ for enforcing semantically similar patches could have the similar dense targets.
Simultaneously, we symmetrically apply context aggregation head $h_\theta(\cdot)$ to produce corresponding context $\mathbf{s}_x^{\mathtt{LAT}}$ from context representations of unmasked patches as:
\begin{equation}
\begin{aligned}
    \mathbf{s}_i^\mathtt{LAT} = h_{\tilde{\theta}}(\{\mathbf{x}_j\}_{j\in\mathcal{P}_i}, \mathbf{x}_t), \quad \mathbf{s}_x^\mathtt{LAT} = h_{{\theta}}(\mathbf{s}_x, \mathbf{x}_c), \\
\end{aligned}
\end{equation}
where $\mathbf{s}_x$ denotes context embeddings and $\mathbf{x}_t, \mathbf{x}_c$ denote the averaged embeddings from all patches in the target encoder and only unmasked patches in the context encoder, respectively.
The cross-attention operator $h_\theta(\cdot)$ and $h_{\tilde{\theta}}(\cdot)$ is formulated as:
\begin{equation}
\begin{aligned}
    h_{\tilde{\theta}}(\{\mathbf{x}_j\}_{j\in\mathcal{P}_i}, \mathbf{x}_t) &= \mathtt{Softmax}\left(\dfrac{\mathbf{x}_t\{\mathbf{x}_j\}_{j\in\mathcal{P}_i}^\top}{\sqrt{D}}\right)\{\mathbf{x}_j\}_{j\in\mathcal{P}_i},\\
    h_{{\theta}}(\mathbf{s}_x, \mathbf{x}_c) &= \mathtt{Softmax}\left(\dfrac{\mathbf{x}_c\mathbf{s}_x^\top}{\sqrt{D}}\right)\mathbf{s}_x, \\
\end{aligned}
\end{equation}
where $D$ is the dimension of embeddings.
For a given target block $\mathbf{s}^\mathtt{LAT}_{y_i}$ corresponding to a target mask $\mathcal{B}_i$, the predictor $g_\theta(\cdot, \cdot)$ takes as input the output of the context patch aggregation head $\mathbf{s}_x^a$ and a mask token for each patch to predict $\{\mathbf{m}_j\}_{j\in\mathcal{B}_i}$, and outputs a dense prediction $\{\hat{\mathbf{s}}_{y_j}^\mathtt{LAT}\}_{j\in\mathcal{B}_i}=g_\theta(\mathbf{s}_x^\mathtt{LAT},\{\mathbf{m}_j\}_{j\in\mathcal{B}_i})$.
The new masking objective is optimized by the average $L_2$ distance between the predicted dense representations $\hat{\mathbf{s}}_{y_j}^\mathtt{LAT}$ and the target dense representation $\mathbf{s}_i$ in the $i$-th block, which is formulated as:
\begin{equation}
    \mathcal{L}_{\mathtt{DMT}\text{-}\mathtt{JEPA}} = \dfrac{1}{|M|}\sum_{i=1}^M \sum_{j\in \mathcal{B}_i} ||\mathbf{s}_i^\mathtt{LAT} - \hat{\mathbf{s}}_{y_j}^\mathtt{LAT}||_2^2,
\end{equation}
where $|M|$ denotes the total number of target blocks, and $\mathcal{B}_i$ is the target mask corresponding to the $i$-th target block.
In terms of semantic similarity among patches, the closer the final target representations, pre-training through these targets would promote the enhancement of learned embeddings that encompass local semantics.

\section{Experiments}

\begin{table*}[t]
	\renewcommand\tabcolsep{6.0pt}
	\centering
        \caption{\textbf{ADE20K semantic segmentation, COCO object detection, and instance segmentation.} We fine-tuned pre-trained ViT-B/16 models to perform ADE20K semantic segmentation and COCO object detection and instance segmentation. The mIoU, AP$^{\mathtt{box}}$, and AP$^{\mathtt{mask}}$ metrics denote the results of ADE20K segmentation, COCO detection, and segmentation, respectively.
        The best are indicated in {\bf bold}.}
   \label{tab: exp_sota_seg_det}
   \vspace{0.05in}
	\scalebox{0.9}{
		\begin{tabular}{llcccc}
			\toprule
			Method & Pre-train data & mIoU & AP$^{\mathtt{box}}$ & AP$^{\mathtt{mask}}$ \\
   \midrule
    Supervised & ImageNet-1K w/ labels & 47.4 & 47.9 & 42.9 \\  
    DINO~\citep{caron2021emerging} & ImageNet-1K & 46.8 & 50.1 & 43.4 \\
    MoCo v3~\citep{chen2021mocov3} & ImageNet-1K & 47.3 & 47.9 & 42.7 \\
    BEiT~\citep{bao2021beit} & ImageNet-1K+DALLE & 47.1 & 49.8 & 44.4 \\
    MAE~\citep{he2021masked} & ImageNet-1K & 48.1 & 50.3 & 44.9 \\ \midrule
    I-JEPA~\citep{assran2023self} & ImageNet-1K & 47.6 & 49.9 & 44.5 \\
    DMT-JEPA (ours) & ImageNet-1K & \bf 49.0 & \bf 50.9 & \bf 45.6 \\
   \bottomrule
    \end{tabular}}
    \vspace{-0.5em}
\end{table*}

\subsection{Experimental setup}

\noindent\textbf{Datasets.}
Following previous methods~\citep{he2021masked,assran2023self}, we use ImageNet-1K~\citep{imagenet_cvpr09} for image classification, MS-COCO~\citep{lin2014coco} for object detection and instance segmentation, and ADE20K~\citep{zhou2017scene,Zhou2018SemanticUO} for semantic segmentation.
We closely follow previous work~\citep{chen2021mocov3,xie2021self-supervised,caron2021emerging}, and adopt the Mask R-CNN~\citep{he2017mask} as the detector. 
The ViT-Base~\citep{dosovitskiy2021an} backbone weights are initialized with weights pre-trained on ImageNet-1K using our DMT-JEPA.
Following the settings in~\citep{he2021masked,bao2021beit}, we use the UPerNet approach~\citep{xiao2018unified} based on our ImageNet-1K pre-trained ViT-Base for evaluation.
For a fair comparison, we fine-tune the detector with the same learning rate in~\citep{he2021masked,bao2021beit}.
For video object segmentation, we use DAVIS-2017~\cite{ponttuset2017davis} dataset containing 60 training, 30 validation, and 60 testing videos.
For local prediction tasks on Clevr~\citep{johnson2016clevr}, we follow the previous work~\citep{assran2023self} and use Clevr/Count and Clevr/Dist.

\noindent\textbf{Evaluation Metrics.}
We follow previous masked image modeling work~\citep{he2021masked,bao2021beit} to report the classification accuracy of linear probing and fine-tuning. 
For object detection and instance segmentation on MS-COCO, we apply AP$^{\mathtt{box}}$ and AP$^{\mathtt{mask}}$ as metrics for the bounding boxes and the instance masks.
mIoU results are reported to evaluate semantic segmentation on ADE20K. 
For video object segmentation on DAVIS-2017, we use Jabri-based $(\mathcal{J} \& \mathcal{F})_m$, $\mathcal{J}_m$, $\mathcal{F}_m$ as metrics to evaluate the quality of frozen representations of image patches by segmenting scenes with the nearest neighbor between consecutive frames.
For local prediction tasks on Clevr, we use object counting and depth prediction to evaluate the linear probing performance of our model.

\begin{table}[t]
	\renewcommand\tabcolsep{6.0pt}
	\centering
         \caption{\textbf{DAVIS video object segmentation.} We perform DAVIS 2017 video object segmentation using ImageNet-1K pre-trained ViT-B/16 and ViT-L/16 models. 
         We report Jabri-based metrics $(\mathcal{J} \& \mathcal{F})_m$, $\mathcal{J}_m$, $\mathcal{F}_m$ to evaluate the quality of pre-trained representations.
         The best results are indicated in {\bf bold}.
         }
           \label{tab: exp_sota_seg_video}
    \vspace{0.05in}
	\scalebox{0.8}{
		\begin{tabular}{lccccc}
			\toprule
			Method & Backbone & $(\mathcal{J} \& \mathcal{F})_m$ & $\mathcal{J}_m$ & $\mathcal{F}_m$  \\
   \midrule
    \multirow{2}{*}{MAE~\citep{he2021masked}} & ViT-B/16 & 51.0 & 49.4 & 52.6 \\
    & ViT-L/16 & 53.4 & 52.5 & 54.3 \\
    \midrule
    \multirow{2}{*}{I-JEPA~\citep{assran2023self}} & ViT-B/16 & 56.2 & 56.1 & 56.3 \\
    & ViT-L/16 & 56.6 & 56.3 & 56.9 \\
    \multirow{2}{*}{DMT-JEPA (ours)} & ViT-B/16 & \textbf{57.7} & \textbf{56.7} & \textbf{58.7} \\
    & ViT-L/16 & \bf 58.3 & \bf 57.3 & \bf 59.2 \\
   \bottomrule
			\end{tabular}}
   \vspace{-1.0em}
\end{table}

\begin{table}[t]
	\renewcommand\tabcolsep{6.0pt}
	\centering
         \caption{\textbf{ImageNet-1K image linear classification.} We perform a linear evaluation on pre-trained ViT-B/16 and ViT-L/16 models for image classification on ImageNet-1K benchmark. We report the top-1 accuracy to evaluate the quality of pre-trained representations. The best results are indicated in {\bf bold}.}
   \label{tab: exp_sota_linprob}
   \vspace{0.05in}
	\scalebox{0.82}{
		\begin{tabular}{lccc}
			\toprule
			Method & Backbone & Epochs & Top-1 
   Acc \\
   \midrule
   data2vec~\citep{baevski2022data2vec} & ViT-L/16 & 1600 & 77.3 \\
   \multirow{2}{*}{MAE~\citep{he2021masked}} & ViT-B/16 & 1600 & 68.0 \\
    & ViT-L/16 & 1600 & 76.0 \\
    \midrule
    \multirow{3}{*}{I-JEPA~\citep{assran2023self}} & ViT-B/16 & 600 & 72.9 \\
    & ViT-L/16 & 600 & 77.5 \\
    & ViT-H/14 & 300 & 79.3 \\
    \multirow{3}{*}{DMT-JEPA (ours)} & ViT-B/16 & 600 &  \textbf{73.8} \\
    & ViT-L/16 & 600 & \bf 78.2 \\
    & ViT-H/14 & 300 & \bf 80.6 \\
   \bottomrule
			\end{tabular}}
    \vspace{-1.0em}
\end{table}

\begin{table*}[!htb]
	\renewcommand\tabcolsep{6.0pt}
	\centering
         \caption{\textbf{ImageNet-1K image fine-tuning classification.} We perform fine-tuning pre-trained ViT-B/16 and ViT-L/16 models for image classification on ImageNet-1K benchmark. 
         We report the top-1 accuracy to evaluate the quality of fine-tuned representations.
         The best results are indicated in {\bf bold}.
         }
           \label{tab: exp_finetune}
    \vspace{0.05in}
	\scalebox{0.9}{
		\begin{tabular}{lccc}
			\toprule
			Method & Pre-train data & Backbone  & Top-1 Accuracy \\
   \midrule
   DINO~\citep{caron2021emerging}  & ImageNet-1K & ViT-B/16 & 82.8 \\
   MAE~\citep{he2021masked} & ImageNet-1K & ViT-B/16 & 83.6 \\ 
    DMT-JEPA (ours) & ImageNet-1K & ViT-B/16 & \textbf{84.6} \\ \midrule
    MAE~\citep{he2021masked} & ImageNet-1K & ViT-L/16 & 85.9 \\ 
    DMT-JEPA (ours) & ImageNet-1K & ViT-L/16 & \textbf{86.6} \\
   \bottomrule
			\end{tabular}}
   \vspace{-0.20in}
\end{table*}

\noindent\textbf{Implementation.}
For input images, the resolution is resized to $224 \times 224$, \textit{i.e.}, $H=W=224$. 
We follow prior work~\citep{he2021masked,assran2023self} and apply a patch size of $16$, \textit{i.e.}, $P=16$.
The small, base, and large models of ViT~\citep{dosovitskiy2021an} architecture are used for experiments.
We set the embedding dimension of the predictor to 384, and keep the number of self-attention heads the same as the backbone context-encoder. 
For the smaller ViT-S/16 and ViT-B/16 context-encoder, we set the depth of the predictor as 6. 
For ViT-L/16 context-encoders, we set the depth of the predictor to 12.
Following I-JEPA~\citep{assran2023self}, we use AdamW to optimize the context-encoder and predictor weights. 
We train our model using the default batch size of 2048, and the learning rate linearly increased from 1e-4 to 1e-3 during the first 15 epochs of pre-training, and decay to 1e-6 following a cosine schedule.
The weight decay is linearly increased from 0.04 to 0.4, and the target-encoder weights are initialized the same as the context-encoder weights, and updated via an exponential moving average.
We use a momentum value of 0.996, and linearly increase this value to 1.0. 
For masking, we use the same strategy and settings as I-JEPA~\citep{assran2023self} for 4 possibly overlapping target blocks masks.
Our small, base, and large models are pre-trained on ImageNet-1K~\citep{imagenet_cvpr09} for 600 epochs.

\subsection{Comparison to prior work}

In this work, we propose a novel and effective discriminative dense target for latent reconstruction-based masked modeling with a joint-embedding predictive architecture.
In order to demonstrate the effectiveness of the proposed DMT-JEPA, we comprehensively compare it to previous mask image modeling baselines~\citep{he2021masked,baevski2022data2vec,chen2022context,assran2023self}.

\noindent\textbf{Detection and Segmentation tasks.}
For the ADE20K semantic segmentation and COCO object detection \& instance segmentation benchmarks, we report the quantitative comparison results in Table~\ref{tab: exp_sota_seg_det}; our method achieved the best results regarding all the metrics compared to previous mask modeling baselines.
In particular, the proposed DMT-JEPA outperforms I-JEPA~\citep{assran2023self}, the current image-based joint-embedding predictive architecture by 0.9@mIoU.
Also, we achieve significant performance gains of 1.0@AP$^{\mathtt{box}}$ and 1.1@AP$^{\mathtt{mask}}$ on COCO object detection and instance segmentation compared to I-JEPA, which indicates the importance of leveraging the
(self-supervised) discriminative dense targets to capture local semantics for dense prediction tasks.
Furthermore, we observed that DMT-JEPA even can achieve better results than the strongest baseline, MAE \citep{he2021masked}, a generative autoencoder architecture for masked image modeling.

In addition, our method shows significant and consistent gains in the DAVIS 2017 video object segmentation benchmark as shown in Table~\ref{tab: exp_sota_seg_video}.
Compared to I-JEPA, ours achieved the results gains of 1.5@$(\mathcal{J} \& \mathcal{F})_m$, 0.6@$\mathcal{J}_m$, and 2.4@$\mathcal{F}_m$ on ViT-B/16.
Moreover, the margins increased more significantly when we evaluated the large-scale backbone, ViT-L/16, by 1.7@$(\mathcal{J} \& \mathcal{F})_m$, which shows a scaling behavior of ours.
Overall, these significant improvements reported in Tables \ref{tab: exp_sota_seg_det} and \ref{tab: exp_sota_seg_video} highlight the superiority of our approach in capturing local semantics during self-supervised pre-training.
This stands in sharp contrast to I-JEPA, which performs even worse than MAE in the same evaluations.

\noindent\textbf{ImageNet linear classification task.}
Here, we validate the quality of our learned global representation by performing the common linear evaluation task on ImageNet-1k. Table~\ref{tab: exp_sota_linprob} summarizes the results; DMT-JEPA outperforms all the baselines in Table~\ref{tab: exp_sota_linprob}. For example, DMT-JEPA achieved 78.2\% top-1 accuracy on ViT-L/16, while MAE and I-JEPA did 76.0\% and 77.5\%, respectively.
These results further indicate the benefit of the proposed method in learning the global semantics of images.

\noindent\textbf{ImageNet fine-tuning comparisons.}
For a comprehensive comparison with DINO and MAE, we follow previous works~\citep{caron2021emerging,he2021masked} and fine-tune pre-trained ViT-B/16 and ViT-L/16 on ImageNet-1K.
Table~\ref{tab: exp_finetune} reports the comparison results with prior approaches using DINO and MAE pre-trained weights.
For instance, our DMT-JEPA outperforms other models with the highest scores of 84.6\% on ViT-B/16 and 86.6\% on ViT-L/16, surpassing DINO and MAE, which achieved scores of 82.8\% and 83.6\% on the ViT-B/16 model, respectively.
These significant improvements underscore our framework's superior capability in grasping the global semantics of images, in contrast to I-JEPA, which has reported finetuning results that are less effective than those achieved by MAE.

\begin{table}[t]
	\renewcommand\tabcolsep{6.0pt}
	\centering
        \caption{\textbf{Clever object counting and depth prediction.} We perform a linear evaluation on pre-trained models for Clever object counting and depth prediction benchmarks. The Clevr/Count and Clevr/Dist metrics denote the result of object counting and depth prediction tasks, respectively. The best results are indicated in {\bf bold}, and the second best ones are \underline{underlined}.}
   \label{tab: exp_sota_lowlevel}
   \vspace{0.05in}
	\scalebox{0.78}{
		\begin{tabular}{lccccccc}
			\toprule
			Method & Backbone & Clevr/Count & Clevr/Dist \\
   \midrule
   DINO~\citep{caron2021emerging} & ViT-B/8 & 86.6 & 53.4\\
   iBOT~\citep{zhou2021ibot} & ViT-L/16 & 85.7 & 62.8 \\ \midrule
    data2vec~\citep{baevski2022data2vec} & ViT-L/16 & 85.3 & 71.3 \\
    \multirow{2}{*}{MAE~\citep{he2021masked}} & ViT-B/16 & \bf 86.6 & \bf 70.8 \\
    & ViT-L/16 & \bf 92.1 & \bf 73.0 \\
    \midrule
    \multirow{3}{*}{I-JEPA~\citep{assran2023self}} & ViT-B/16 & 82.2 & 70.7 \\
    & ViT-L/16 & 85.6 & 71.2 \\
    & ViT-H/14 & 86.7 & 72.4 \\
    \multirow{3}{*}{DMT-JEPA (ours)} & ViT-B/16 & \underline{83.5} & \underline{71.1} \\
    & ViT-L/16 & \underline{87.1} & \underline{71.8} \\
    & ViT-H/14 & \bf 87.9 & \bf 73.1 \\
   \bottomrule
 \end{tabular}}
 \vspace{-0.5em}
\end{table}

\noindent\textbf{Other low-level tasks.}
We also present additional Clevr benchmarks for measuring abilities of object-counting and depth prediction.
Table~\ref{tab: exp_sota_lowlevel} shows linear evaluation results of DMT-JEPA on the Clevr counting and depth benchmarks.
Compared to I-JEPA~\citep{assran2023self}, we achieve the results gains of 1.3@Clevr/Count and 0.4@Clevr/Dist using ViT-B/16.
Moreover, we observe similar scaling behavior in Table~\ref{tab: exp_sota_seg_video}, with increased improvements on ViT-L/16, while our model on ViT-H can outperform MAE.
These results show that JEPA objectives on the embedding space have the potential to degrade dense representation learning and outperform view-invariance compared to pixel reconstruction.

\noindent\textbf{Computational Comparison with I-JEPA.}
In order to comprehensively assess the efficiency of the proposed DMT-JEPA, we compared it with I-JEPA~\citep{assran2023self}, the state-of-the-art image-based joint-embedding predictive architecture, on total pre-training epochs, max memory per GPU, and running time per step in Table~\ref{tab: exp_cost}.
We can observe that our DMT-JEPA achieves comparable computation costs in terms of all metrics, especially on total pre-training epochs and max memory usage.
More importantly, we achieve much better downstream performance regarding segmentation \& detection in Table~\ref{tab: exp_sota_seg_det} \&~\ref{tab: exp_sota_seg_video}, image classification in Table~\ref{tab: exp_sota_linprob} \&~\ref{tab: exp_finetune}, and other low-level tasks in Table~\ref{tab: exp_sota_lowlevel}.
These computational analyses further demonstrate the efficiency of our novel framework.

\begin{table}[t]
	\renewcommand\tabcolsep{6.0pt}
	\centering
        \caption{\textbf{Analysis on computational costs.} We perform computational analyses on pre-trained ViT-B/16 models for comparison with I-JEPA.
        The best results are indicated in {\bf bold}.}
   \label{tab: exp_cost}
   \vspace{0.05in}
	\scalebox{0.75}{
		\begin{tabular}{lccc}
			\toprule
            \multirow{2}{*}{Method} & Pre-train & Max Memory & Running Time \\
            & Epochs & per GPU (GB) & per Step (ms) \\
            \midrule
             I-JEPA~\citep{assran2023self} & 600 & 21.9 & 606.4 \\
             DMT-JEPA (ours) & 600 &  21.9 & 608.2 \\
            \bottomrule
			\end{tabular}}
  \vspace{-0.5em}
\end{table}

\begin{table}[t]
	\renewcommand\tabcolsep{6.0pt}
	\centering
        \caption{\textbf{Ablation studies on component analysis.} We perform ablation studies on Masked Semantic Neighboring (MSN) and Local Aggregation Target (LAT) modules using a pre-trained ViT-S/16 model on the DAVIS benchmark. The best results are indicated in {\bf bold}.
        }
        \vspace{0.05in}
   \label{tab: exp_ablation}
	\scalebox{0.9}{
		\begin{tabular}{ccccc}
			\toprule
            MSN & LAT & $(\mathcal{J} \& \mathcal{F})_m$ & $\mathcal{J}_m$ & $\mathcal{F}_m$ \\
            \midrule
            \xmark & \xmark & 53.7 & 52.5 & 54.8 \\
            \cmark & \xmark & 55.2 & 54.3 & 56.1 \\
            \xmark & \cmark & 54.6 & 53.3 & 55.9 \\
            \cmark & \cmark & \bf 57.1 & \bf 55.7 & \bf 58.5 \\
            \bottomrule
			\end{tabular}}
  \vspace{-0.5em}
\end{table}

\begin{table}[t]
	\renewcommand\tabcolsep{6.0pt}
	\centering
        \caption{\textbf{Ablation studies on local aggregation head.} We perform ablation studies on the cross-attention layers in the Local Aggregation Target module using ViT-B/16 pre-trained models. 
        The best results are indicated in {\bf bold}.}
   \label{tab: exp_pooling}
   \vspace{0.05in}
	\scalebox{0.9}{
		\begin{tabular}{ccccc}
			\toprule
            Local Aggregation Head & $(\mathcal{J} \& \mathcal{F})_m$ & $\mathcal{J}_m$ & $\mathcal{F}_m$ \\
            \midrule
             Average-pooling & 55.2 & 54.3 & 56.1 \\
             Max-pooling & 55.6 & 54.6 & 56.6 \\
             Cross-attention & \bf 57.1 & \bf 55.7 & \bf 58.5 \\
            \bottomrule
			\end{tabular}}
  \vspace{-0.5em}
\end{table}

\begin{figure}[t]
\centering
\includegraphics[width=0.98\linewidth]{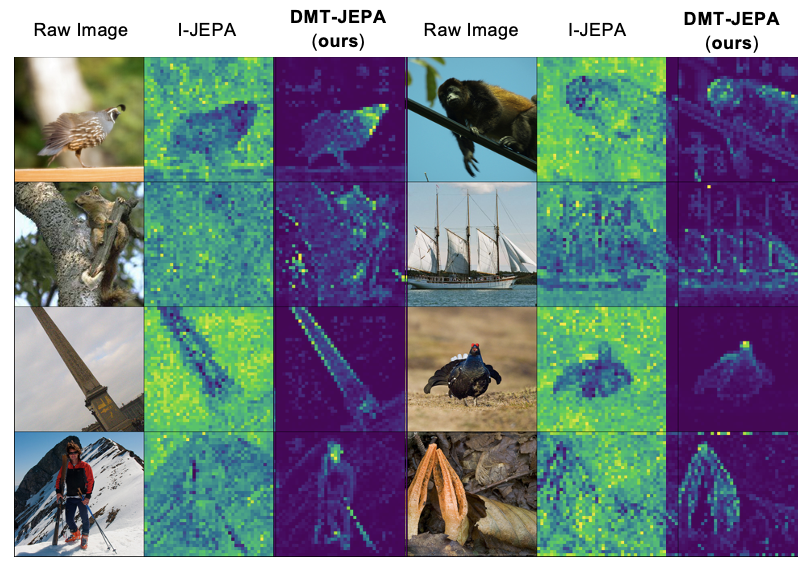}
\vspace{-0.5em}
\caption{{\bf Qualtitative visualization of learned attention maps using ViT-B/16 model. } 
Columns for each sample denote the original image, attention maps from I-JEPA~\citep{assran2023self}, and attention maps from our DMT-JEPA.
Our DMT-JEPA achieves much better attention maps.
}
\vspace{-0.5em}
\label{fig: exp_vis_attention}
\end{figure}

\begin{table}[t]
	\renewcommand\tabcolsep{6.0pt}
	\centering
        \caption{\textbf{Ablation studies on aggregation head.} We perform ablation studies on context and target aggregation heads in Local Aggregation Target modules using two types (Cross-attention \& Self-attention). 
        The best results are indicated in {\bf bold}.}
   \label{tab: exp_head}
   \vspace{0.05in}
	\scalebox{0.85}{
		\begin{tabular}{ccccc}
			\toprule
            Context Head $h_\theta$ & Target Head $h_{\tilde{\theta}}$ & $(\mathcal{J} \& \mathcal{F})_m$ & $\mathcal{J}_m$ & $\mathcal{F}_m$ \\
            \midrule
            Cross-attention & Cross-attention & \bf 57.1 & \bf 55.7 & \bf 58.5 \\
            Self-attention & Cross-attention & 53.7 & 52.5 & 54.9 \\
            Self-attention & Self-attention & 50.6 & 49.5 & 51.7 \\
            \bottomrule
			\end{tabular}}
  \vspace{-0.5em}
\end{table}

\subsection{Experimental Analysis}

In this section, we performed ablation studies to demonstrate the benefit of the proposed Masked Semantic Neighboring and Local Aggregation Target modules.
Here, we conducted extensive experiments on ImageNet-1k pre-trained ViT-S/16 to explore the impact of cross-attention and pooling, types of local aggregation heads, and learned meaningful patch-level representations.

\noindent\textbf{Masked Semantic Neighboring \& Local Aggregation Target.}
In order to demonstrate the effectiveness of Masked Semantic Neighboring (MSN) and Local Aggregation Target (LAT), we ablate the impact of each module and report the quantitative results on DAVIS 2017 video object segmentation benchmark with ViT-S/16 in Table~\ref{tab: exp_ablation}.
As shown in the table, adding MSN to the vanilla baseline highly increases the results of 1.5@$(\mathcal{J} \& \mathcal{F})_m$, which validates the benefit of masked semantic neighboring in finding semantically similar neighboring patches for masked patches.
Meanwhile, introducing only LAT in the baseline increases the video segmentation performance regarding all metrics.
More importantly, incorporating MSN and LAT into the baseline significantly raises the performance by 3.4@$(\mathcal{J} \& \mathcal{F})_m$.
These improving results validate the importance of MSN and LAT in extracting local semantics from (self-supervised) discriminative dense targets for better representations.

\noindent\textbf{Ablation on Cross-attention vs Pooling in Local Aggregation Head.}
To validate the effectiveness of using cross-attention layers, we ablated the layer using average-pooling and max-pooling operators.
The comparison results are reported in Table~\ref{tab: exp_pooling}.
As can be observed, replacing the cross-attention layer with average-pooling deteriorates the results by 1.9@$(\mathcal{J} \& \mathcal{F})_m$, 1.4@$\mathcal{J}_m$, and 2.4@$\mathcal{F}_m$.
This might be because average pooling leads to collapsing and losing discrepancy across patch tokens.
Our key is each query patch has distinct neighborhood patches, and objectively derived from the distinct neighborhoods will ensure their distinct target objectives.
Meanwhile, using the max-pooling operator highly decreases the results in terms of all metrics. 
These results validate the importance of cross-attention layers in preventing losing discrepancy for distinct target representations during self-supervised pre-training.

\noindent\textbf{Types of Local Aggregation Heads.}
Local aggregation heads affect the ability of the proposed method to aggregate dense targets and context with local semantics.
To explore such effects more comprehensively, we varied the type of Local Aggregation Heads from cross-attention and self-attention operators asymmetrically.
We report the comparison results on the DAVIS benchmark with ViT-S/16 in Table~\ref{tab: exp_head}.
When both the context and target head use the cross-attention operators, we achieve the best performance in terms of all metrics.
Replacing cross-attention operators with self-attention operators significantly worsens the results in terms of all metrics.
These results indicate the difficulty in the asymmetric use of cross-attention aggregation heads, 
as the target aggregation head is updated using an exponential moving average of the context head weights, and it cannot be solely trained on its architecture.

\noindent\textbf{Visualizations of Learned Attention Maps.}
Learning discriminative attention maps is one of the key aspects
of capturing local semantics for downstream tasks of dense prediction type, such as segmentation and detection.
To better evaluate the quality of learned attention maps, we visualize the averaged maps from different heads in the last attention layer by using a pre-trained ViT-B/16 target encoder.
For a more comprehensive comparison, we also add the attention maps from I-JEPA~\citep{assran2023self} target encoder.
The qualitative visualization results are shown in Figure~\ref{fig: exp_vis_attention}.
Note that columns for each image sample represent the original image, attention maps from the target encoder in I-JEPA, and attention maps from the target encoder in our DMT-JEPA. 
As can be seen, attention maps from the target encoder in our DMT-JEPA are discriminative and focus on the object semantics in the image.
However, the attention maps from the target encoder in I-JEPA activate both the foreground and background and can not effectively discriminate the object semantics, because they did not incorporate local semantics into target representations as our DMT-JEPA did.
Meanwhile, the attention maps in our DMT-JEPA have more focus on the details of foreground objects than that from the target encoder, indicating the effectiveness of our local aggregation target module in generating target representations with local semantics.
These high-quality visualization results further demonstrate the superiority of our new framework in learning meaningful representations with local semantics during self-supervised training, compared to I-JEPA, the state-of-the-art image-based joint-embedding predictive architecture.

\section{Related Work}

\noindent{\textbf{Self-supervised Visual Representation Learning.}}
Contrastive learning has been widely used in self-supervised vision transformers~\citep{chen2021mocov3,xie2021self-supervised,caron2021emerging,yun2022patch,mo2023mcvt} to achieve promising performance on visual downstream tasks. 
Typically, MoCo-v3~\citep{chen2021mocov3} introduced a momentum encoder in ViT~\citep{dosovitskiy2021an} to minimize the distance between representations of two augmented views from the base encoder and momentum one.
To capture the local-to-global alignment, DINO~\citep{caron2021emerging} used a momentum encoder with multi-crop training to achieve knowledge distillation in the vision transformer. 
Recently, SelfPatch~\citep{yun2022patch} utilized neighboring information to enhance self-supervision, a strategy that proved beneficial for dense representation learning. 
While we draw inspiration from the concept of leveraging neighboring information, DMT-JEPA uniquely focuses on predicting information about neighboring patches without incorporating a global component, such as DINO~\citep{caron2021emerging} loss, in stark contrast to SelfPatch, which heavily relies on global supervision. 
This distinct approach significantly enhances DMT-JEPA's scalability, enabling it to achieve significant improvements in image classification tasks with larger models, thereby surpassing the benefits offered by SelfPatch.

\noindent{\textbf{Masked Image Modeling.}}
Masked image modeling has been extensively explored in previous literature~\citep{bao2021beit,atito2021sit,he2021masked,wei2022masked,xie2022SimMIM} to reconstruct the masked image patch given the unmasked counterpart as clues. 
Early approaches~\citep{bao2021beit,atito2021sit,he2021masked,li2021mst,shi2022adversarial} designed customized masking strategies (\textit{e.g.}, random, block-wise) as pre-text tasks during pre-training.
Given features extracted from the 25\% unmasked patches, the seminal work, MAE~\citep{he2021masked} directly reconstructed missing pixels of 75\% masked patches, and showed promising performances on various downstream tasks.

\noindent{\textbf{Joint-Embedding Predictive Architectures.}}
Joint-Embedding Predictive Architectures (JEPA;~\cite{lecun2022path}) learn to predict the embeddings of a signal from a compatible signal using a predictor conditioned on a latent variable to achieve prediction.
Based on JEPA, the image-based JEPA, I-JEPA~\citep{assran2023self}, recently proposed to predict multiple target block representations given the output of the context encoder under a masking strategy.
The key characteristic of I-JEPA is that its objective is located on the embedding space, while masked image modeling models do on the pixel (or token) space.
Despite it enabling efficient pre-training of I-JEPA, however, the masked modeling target on the embedding space could pose degradation in various downstream tasks if the target representations were non-discriminative or 
lack local semantics among themselves.
In contrast, we aim to develop a novel masked modeling objective incorporated with JEPA to generate discriminative target representations capturing local semantics for enhanced masked modeling on the embedding space.

\section{Conclusion}

We introduce a novel discriminative embedding objective tailored for masked modeling on the joint-embedding predictive architecture from unlabeled images.
To tackle this, we aim to produce semantically meaningful target representations in a self-supervised manner by leveraging the prior that neighboring patches often contain similar semantics.
To be specific, we first search semantically similar patches for a masked patch within its neighborhood by computing their similarities on the representation space.
Then we generate the aggregated representations from the selected neighboring patches to serve as a masked modeling objective via a lightweight cross-attention head.
Finally, the proposed objective would accelerate learned representations of semantically similar patches being closer, which can be advantageous in understanding local semantics within images.
Through extensive experiments, we have demonstrated our models are not only effective in dense prediction types of downstream tasks but also show strong performance in image classification tasks.
We believe that our work would highlight the effectiveness of considering a discriminative target for masked modeling on the embedding space.

\newpage
\section*{Impact Statement}

We believe that mask reconstruction on latent space will gain much attention on its efficacy and efficiency, so the study of training objectives on latent space is crucial. 
Therefore, our investigation of the latent objective is invaluable and could inspire future research direction to explore better SSL objectives, unlike the previous stereotype of masked reconstruction objective on pixel space.


\bibliography{reference}
\bibliographystyle{icml2024}

\newpage

\appendix

\counterwithin{figure}{section}
\counterwithin{table}{section}
\counterwithin{algorithm}{section}

\section*{Appendix}

In this appendix, we provide the following materials:
\begin{itemize}
    \item Additional experiments on segmentation comparison with DINOv2~\citep{oquab2023dinov2} in Section~\ref{sec:appendix_exp},
    \item Additional analyses on the effectiveness of cross-attention layers in local aggregation head, and computational comparison with I-JEPA~\citep{assran2023self} in Section~\ref{sec:appendix_analy},
    \item Additional visualizations on the cosine similarity maps of query patches and learned attention maps in Section~\ref{sec:appendix_vis}.
\end{itemize}

\section{Additional Experiments}\label{sec:appendix_exp}

In order to further demonstrate the effectiveness of the proposed DMT-JEPA in learning local semantics during self-supervised pre-training, we conduct experiments on segmentation comparison on ADE20K with DINOv2~\citep{oquab2023dinov2}, and extensive experiments for standardization comparisons with more baseline methods.

\noindent\textbf{Segmentation Comparison with DINOv2.}
For a more comprehensive comparison with DINOv2~\citep{oquab2023dinov2}, the recent strong self-supervised baseline trained on a large-scale dataset, we compare our ImageNet pre-trained ViT-B/16 models with Figure 6 in the original DINOv2 paper using ImageNet pre-trained ViT-L/16 models on ADE20K semantic segmentation.
The comparison results are reported in Table~\ref{tab: exp_seg_dinov2}.
We can observe that our DMT-JEPA with ViT-B/16 backbone achieves better results than DINOv2 using ViT-L/16 using different pre-training resolutions.
However, it should be noted that DINOv2~\citep{oquab2023dinov2} performed linear evaluation settings on the ADE20K dataset by freezing representations at different resolutions.
Although the comparison is not totally fair, these results further demonstrate the competitive fine-tuning performance on segmentation by learning local semantics during self-supervised pre-training. 
We believe that our new framework can be scaled up to more pre-training data to achieve better downstream performances.

\begin{table*}[!htb]
	\renewcommand\tabcolsep{6.0pt}
	\centering
        \caption{\textbf{ADE20K semantic segmentation.} We fine-tuned pre-trained ViT-B/16 models to perform ADE20K semantic segmentation.
        Note that DINOv2~\citep{oquab2023dinov2} performed linear evaluation settings on the ADE20K benchmark by freezing representations at different resolutions.
        The best results are indicated in {\bf bold}.}
   \label{tab: exp_seg_dinov2}
   \vspace{0.05in}
	\scalebox{0.9}{
		\begin{tabular}{llccc}
			\toprule
			Method & Pre-train data & Backbone & Pre-train Resolution & mIoU \\
   \midrule
    \multirow{2}{*}{DINOv2~\citep{oquab2023dinov2}} & ImageNet-1K &  ViT-L/16 & 224 & 43.0 \\
    & ImageNet-1K &  ViT-L/16 & 416 & 46.2 \\ \midrule
    DMT-JEPA (ours) & ImageNet-1K & ViT-B/16 & 224 & \bf 49.0 \\
   \bottomrule
    \end{tabular}}
    \vspace{-0.05in}
\end{table*}

\begin{table*}[!htb]
	\renewcommand\tabcolsep{6.0pt}
	\centering
         \caption{\textbf{Benchmarks with mask modeling methods.} 
         We perform downstream tasks on ImageNet-1K pre-trained ViT-B/16 models on diverse benchmarks to evaluate the quality of fine-tuned representations.
         The best results are indicated in {\bf bold}.
         }
           \label{tab: exp_benchmark_mlm}
    \vspace{0.05in}
	\scalebox{0.9}{
		\begin{tabular}{lccccccc}
\toprule
Method & mIoU & AP$^{\mathtt{box}}$ & AP$^{\mathtt{mask}}$ & Clevr/Count & Clevr/Dist & Linear Prob & Finetune \\ \midrule
BEiT~\cite{bao2021beit}            & 47.1          & 49.8             & 44.4              & 82.5                 & 70.2               & 56.7                & 83.4              \\ 
MAE~\cite{he2021masked}             & 48.1          & 50.3             & 44.9              & \textbf{86.6}        & 70.8               & 68.0                & 83.6              \\
iBOT~\cite{zhou2021ibot}            & \textbf{50.0} & \textbf{51.2}    & 44.2              & 81.2                 & 70.3               & \textbf{79.5}       & 84.0              \\ 
I-JEPA~\cite{assran2023self}          & 47.6          & 49.9             & 44.5              & 82.2                 & 70.7               & 72.9                & 83.5              \\ 
DMT-JEPA (ours) & 49.0          & 50.9             & \textbf{45.6}     & 83.5                 & \textbf{71.1}      & 73.8                & \textbf{84.6}     \\ \bottomrule
			\end{tabular}}
   \vspace{-0.05in}
\end{table*}

\begin{table*}[!htb]
	\renewcommand\tabcolsep{6.0pt}
	\centering
         \caption{\textbf{Benchmarks with distillation modeling methods.} 
         We perform downstream tasks on ImageNet-1K pre-trained ViT-B/16 models on diverse benchmarks to evaluate the quality of fine-tuned representations.
         The best results are indicated in {\bf bold}.
         }
           \label{tab: exp_benchmark_distillation}
    \vspace{0.05in}
	\scalebox{0.9}{
		\begin{tabular}{lccccc}
\toprule
Method & mIoU & AP$^{\mathtt{box}}$ & AP$^{\mathtt{mask}}$ & Linear Prob & Finetune \\ \midrule
DTM~\cite{kim2024morphing}              & 53.1         & -                & -                 & 77.6                & 85.4             \\ 
dBOT~\cite{liu2023exploring}            & 49.5         & 52.7             & 45.7              & -                   & 84.5             \\ \hline
Hybrid Distill~\cite{shi2023hybrid} & \textbf{49.1}& 50.3             & 44.2              & -                   & 83.7             \\ 
DMT-JEPA (ours)     & 49.0         & \textbf{50.9}    & \textbf{45.6}     & \bf 73.8                & \textbf{84.6}    \\ \bottomrule
			\end{tabular}}
   \vspace{-0.05in}
\end{table*}

\noindent\textbf{Standardization of Baseline Methods.}
In Table~\ref{tab: exp_benchmark_mlm}, we expanded our experimental results using ViT-B/16 to include direct comparisons with both iBOT~\cite{zhou2021ibot} and BEiT~\cite{bao2021beit} across all relevant tasks, thus providing a more consistent evaluation framework and demonstrating the versatility of DMT-JEPA.
We also included detailed comparisons with DTM, dBOT, and Hybrid Distill using ViT-B/16 for a more comprehensive evaluation. 
The quantitative results are reported in Table~\ref{tab: exp_benchmark_distillation}. 
Note that both DTM~\cite{kim2024morphing} and dBOT~\cite{liu2023exploring} use pre-trained CLIP as a teacher to train the backbone ViT-B/16 on ImageNet-1K, which makes the comparison fully fair. 
However, our DMT-JEPA outperforms Hybrid Distill~\cite{shi2023hybrid} without using CLIP regarding most downstream tasks, especially on fine-tuning and object detection.

\section{Additional Analysis}\label{sec:appendix_analy}

In this section, we performed ablation studies to demonstrate the impact of neighbors and dense pairs in the proposed context and target aggregation heads in Local Aggregation Target modules.
Our ablation experiments are based on ImageNet-1k pre-trained ViT-S/16 models.

To explore the impact of neighbors in neighboring and the number of selected dense pairs, we ablated the neighbors from $\{3\times 3, 5\times 5$, All patches$\}$ and varied the number of dense pairs from $\{1,2,4,8\}$.
The quantitative results on the DAVIS benchmark with ViT-S/16 are reported in Table~\ref{tab: exp_neb_positive}.
As shown in the table, the proposed DMT-JEPA achieved the best performance of $(\mathcal{J} \& \mathcal{F})_m$ when we use $3\times 3$ neighbors and $4$ dense pairs.
With the increased number of dense pairs from $1$ to $4$, the proposed method consistently increases performance as more semantically similar target pairs are extracted.
Nevertheless, increasing the number of dense pairs from 4 to 8 will not continually improve the results since $4$ dense pairs might be enough to extract the learned dense representations using ViT-S/16.
Furthermore, replacing $3\times 3$ neighbors with all patches significantly deteriorates the performance of all metrics. 
These results indicate the importance of selecting semantically meaningful neighboring patches for capturing local semantics.

\begin{table*}[t]
	\renewcommand\tabcolsep{6.0pt}
	\centering
        \caption{\textbf{Ablation studies on hyperparameters.} We perform ablation studies using a pre-trained ViT-S/16 model to explore effects on neighbors and dense pairs in the Masked Semantic Neighboring module. The best results are indicated in {\bf bold}.
        }
   \label{tab: exp_neb_positive}
   \vspace{0.05in}
	\scalebox{0.9}{
		\begin{tabular}{ccccc}
			\toprule
            Neighbors & \# Dense Pairs & $(\mathcal{J} \& \mathcal{F})_m$ & $\mathcal{J}_m$ & $\mathcal{F}_m$ \\
            \midrule
            3$\times$3 & \multirow{3}{*}{4} & \bf 57.1 & \bf 55.7 & \bf 58.5 \\
            5$\times$5  & & 56.3 & 54.9 & 57.7 \\
            All patches  & & 48.5 & 46.7 & 50.3 \\ \midrule
            \multirow{3}{*}{3$\times$3} & 1 & 55.9 & 54.3 & 57.5 \\
            & 2 & 56.3 & 54.9 & 57.7 \\
            & 8 & 56.7 & 55.3 & 58.1 \\
            \bottomrule
			\end{tabular}}
  \vspace{-0.05in}
\end{table*}

\section{Additional Visualizations}\label{sec:appendix_vis}

\begin{figure*}[!htb]
\centering
\includegraphics[width=0.98\linewidth]{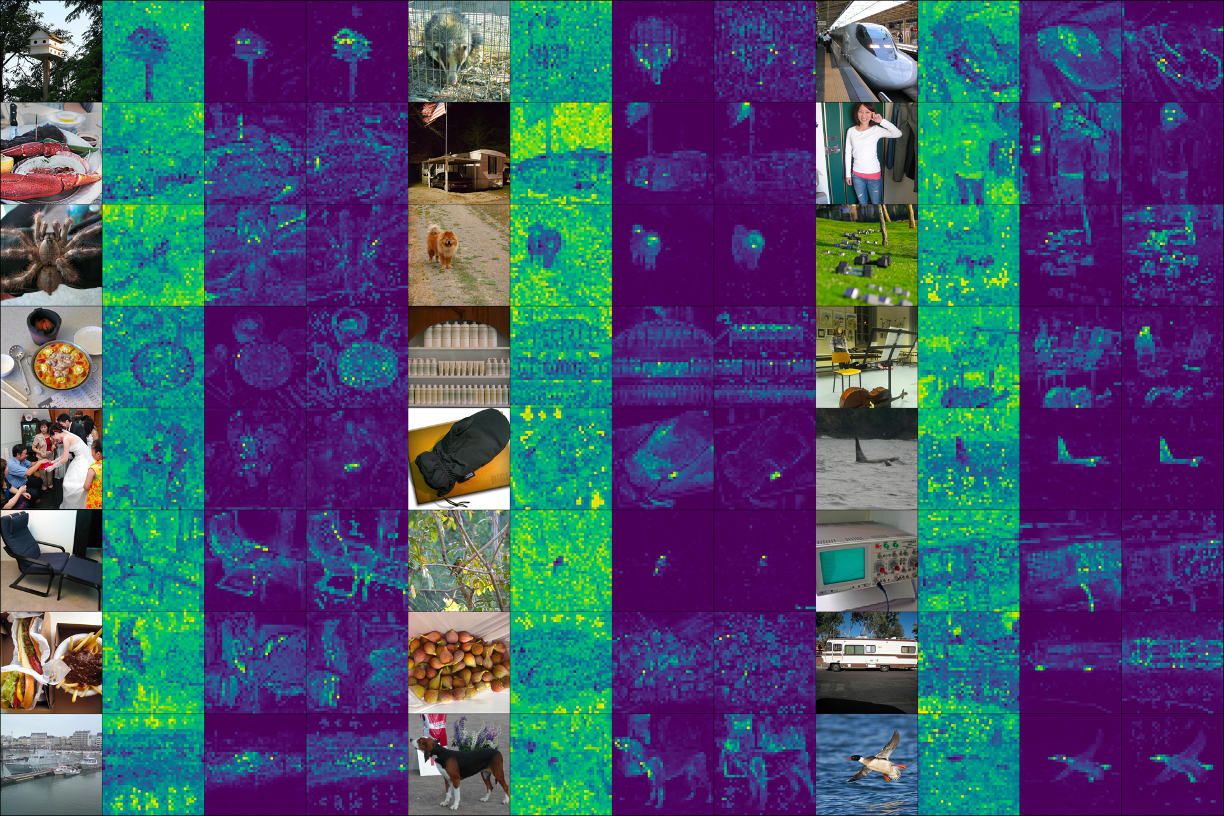}
\caption{Qualtitative visualization of learned attention maps using ViT-B/16 model. Columns for each sample denote the original image, attention maps from target encoder in I-JEPA~\citep{assran2023self},  attention maps from target encoder in our DMT-JEPA, and attention maps from the local aggregation head in our DMT-JEPA.
Our DMT-JEPA achieves much better attention maps.
}
\label{fig: exp_vis_attention1}
\end{figure*}

\begin{figure*}[t]
\centering
\includegraphics[width=0.98\linewidth]{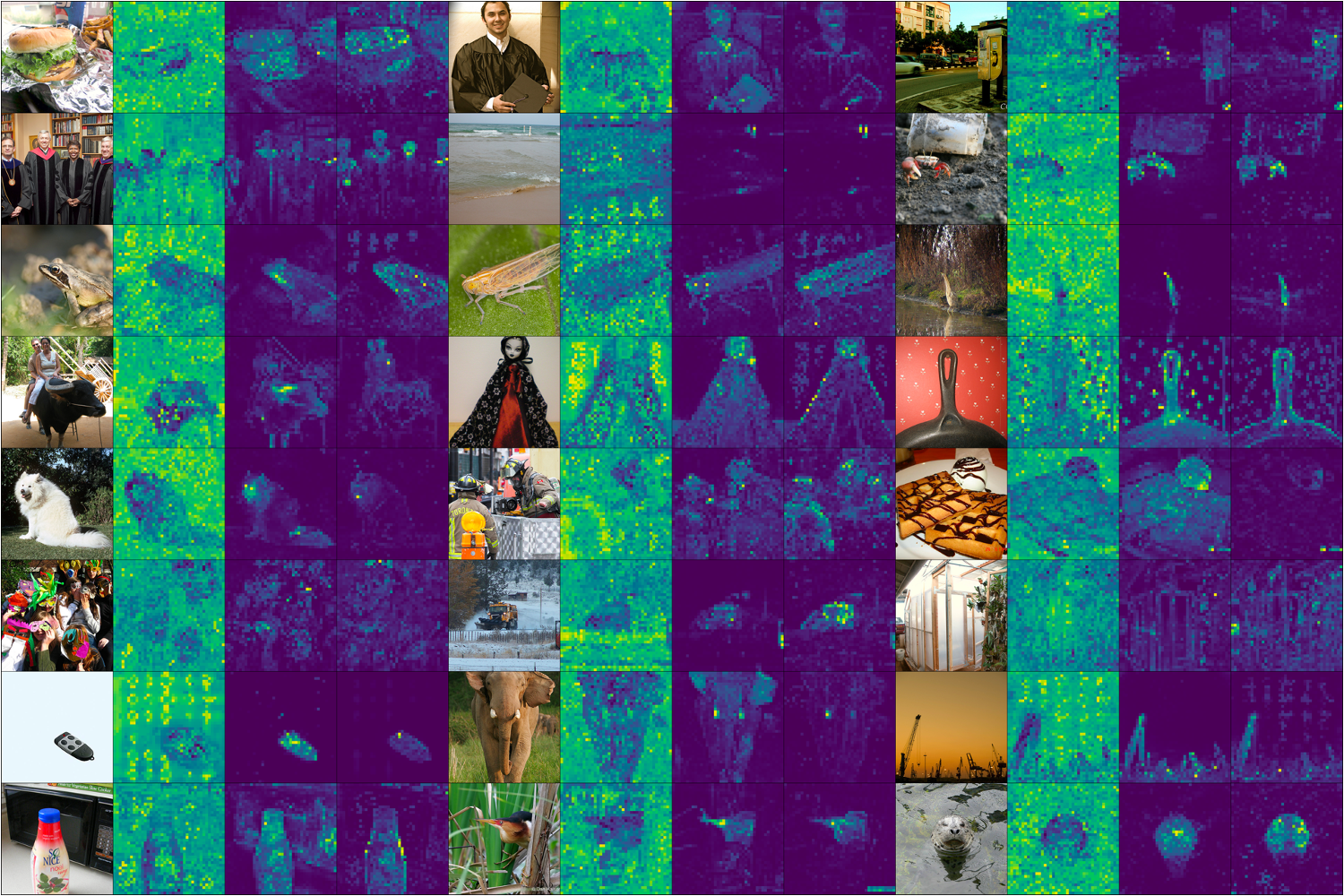}
\caption{Qualtitative visualization of learned attention maps using ViT-B/16 model. Columns for each sample denote the original image, attention maps from target encoder in I-JEPA~\citep{assran2023self},  attention maps from target encoder in our DMT-JEPA, and attention maps from the local aggregation head in our DMT-JEPA.
Our DMT-JEPA achieves much better attention maps.
}
\label{fig: exp_vis_attention2}
\end{figure*}

\begin{figure*}[t]
\centering
\includegraphics[width=0.98\linewidth]{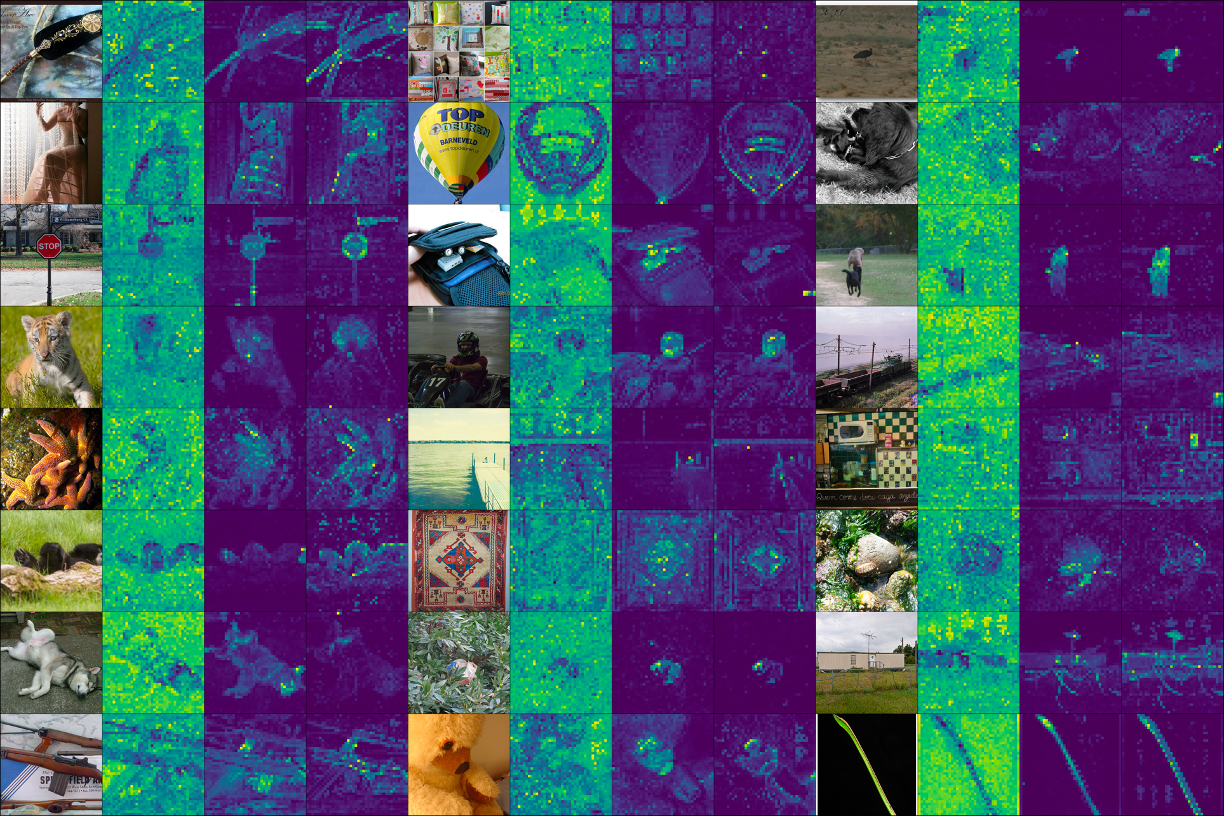}
\caption{Qualtitative visualization of learned attention maps using ViT-B/16 model. Columns for each sample denote the original image, attention maps from target encoder in I-JEPA~\citep{assran2023self},  attention maps from target encoder in our DMT-JEPA, and attention maps from the local aggregation head in our DMT-JEPA.
Our DMT-JEPA achieves much better attention maps.
}
\label{fig: exp_vis_attention3}
\end{figure*}

\begin{figure*}[!htb]
\centering
\includegraphics[width=0.9\linewidth]{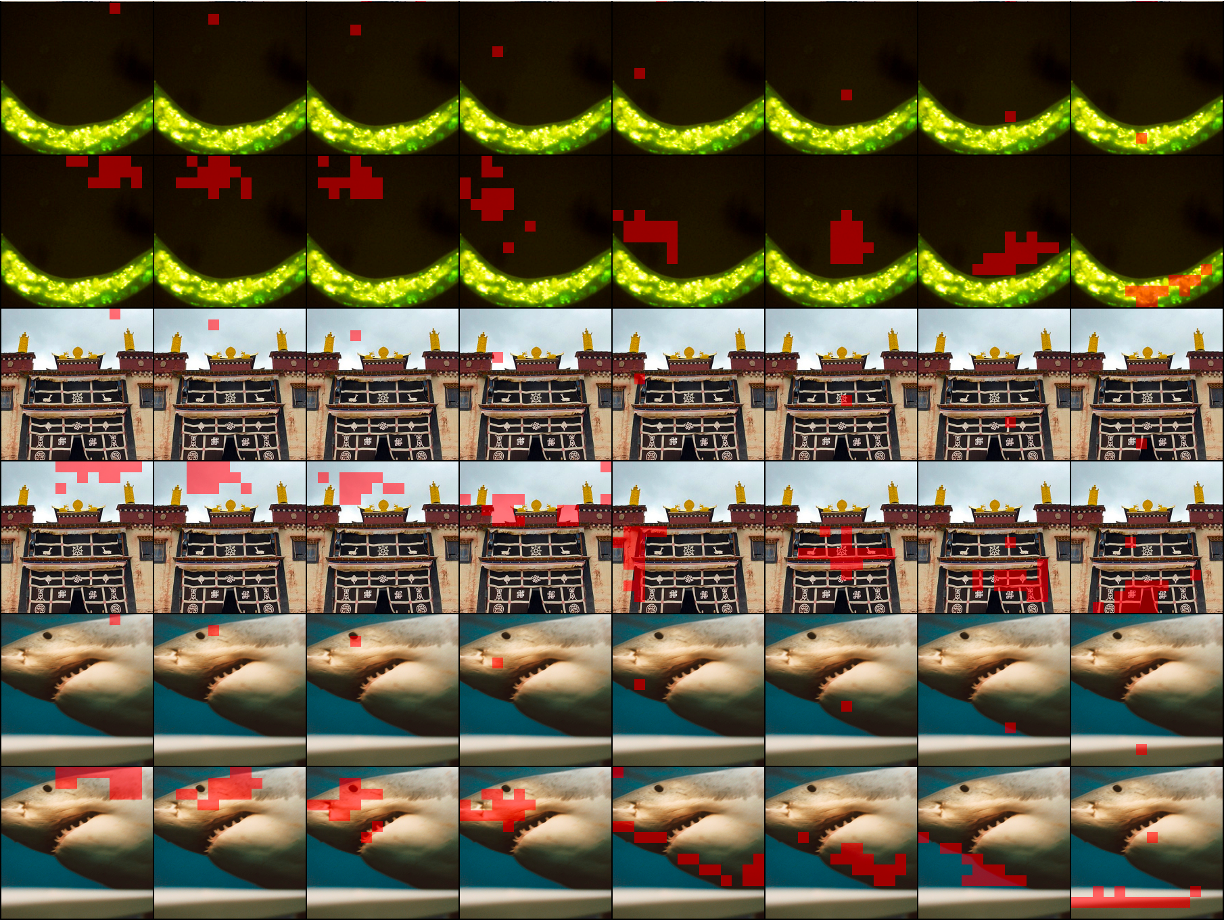}
\caption{Qualtitative visualization of learned cosine similarity maps given query patches using ViT-B/16. Rows for each sample denote the location of the given query patch and the top-10\% patches. Our DMT-JEPA performs effectively by encouraging the model to learn local semantics.
}
\label{fig: exp_vis_sim_one}
\end{figure*}

\begin{figure*}[t]
\centering
\includegraphics[width=0.9\linewidth]{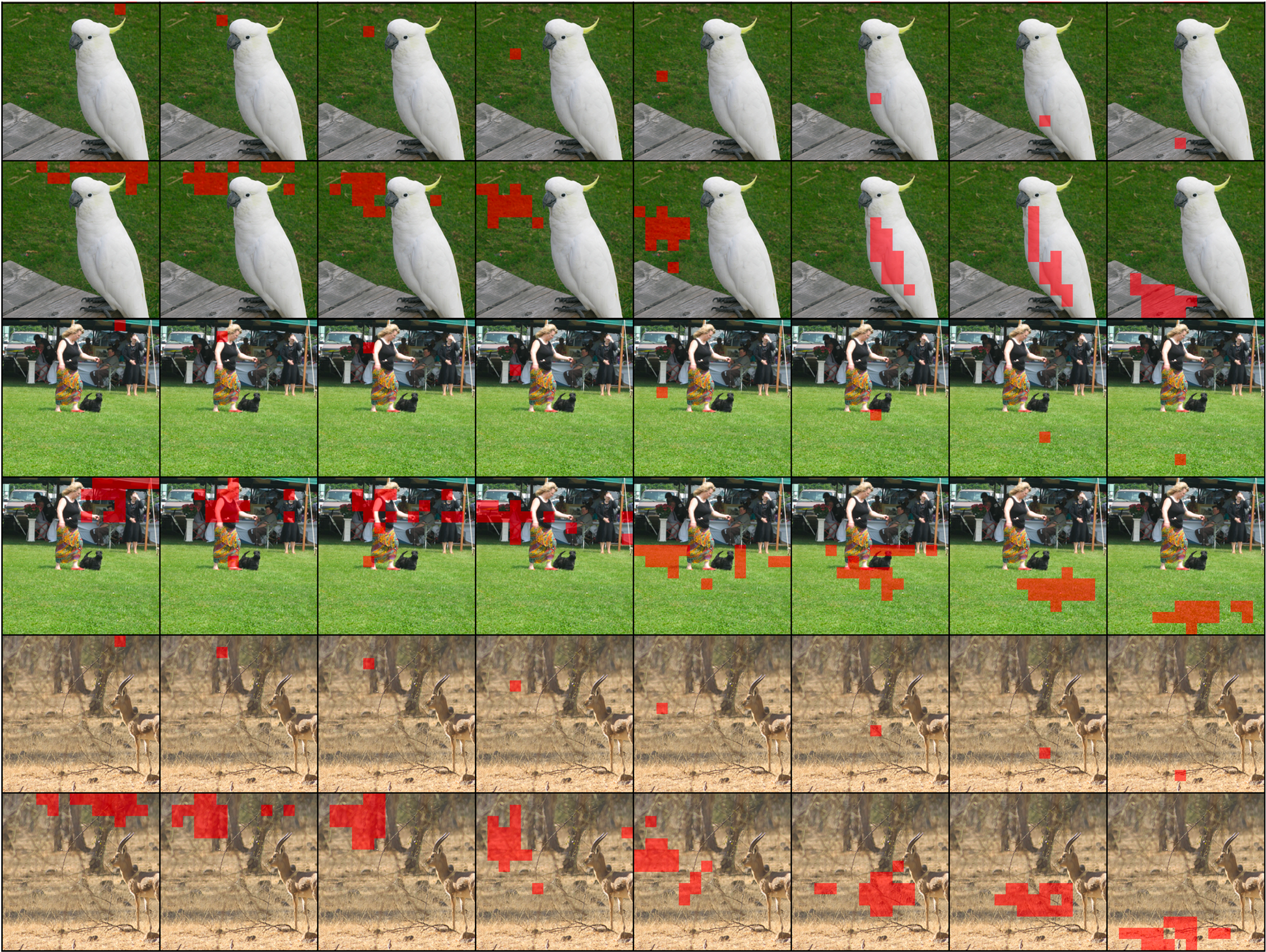}
\caption{Qualtitative visualization of learned cosine similarity maps given query patches using ViT-B/16. Rows for each sample denote the location of the given query patch and the top-10\% patches. Our DMT-JEPA performs effectively by encouraging the model to learn local semantics.
}
\label{fig: exp_vis_sim_two}
\end{figure*}

\begin{figure*}[t]
\centering
\includegraphics[width=0.9\linewidth]{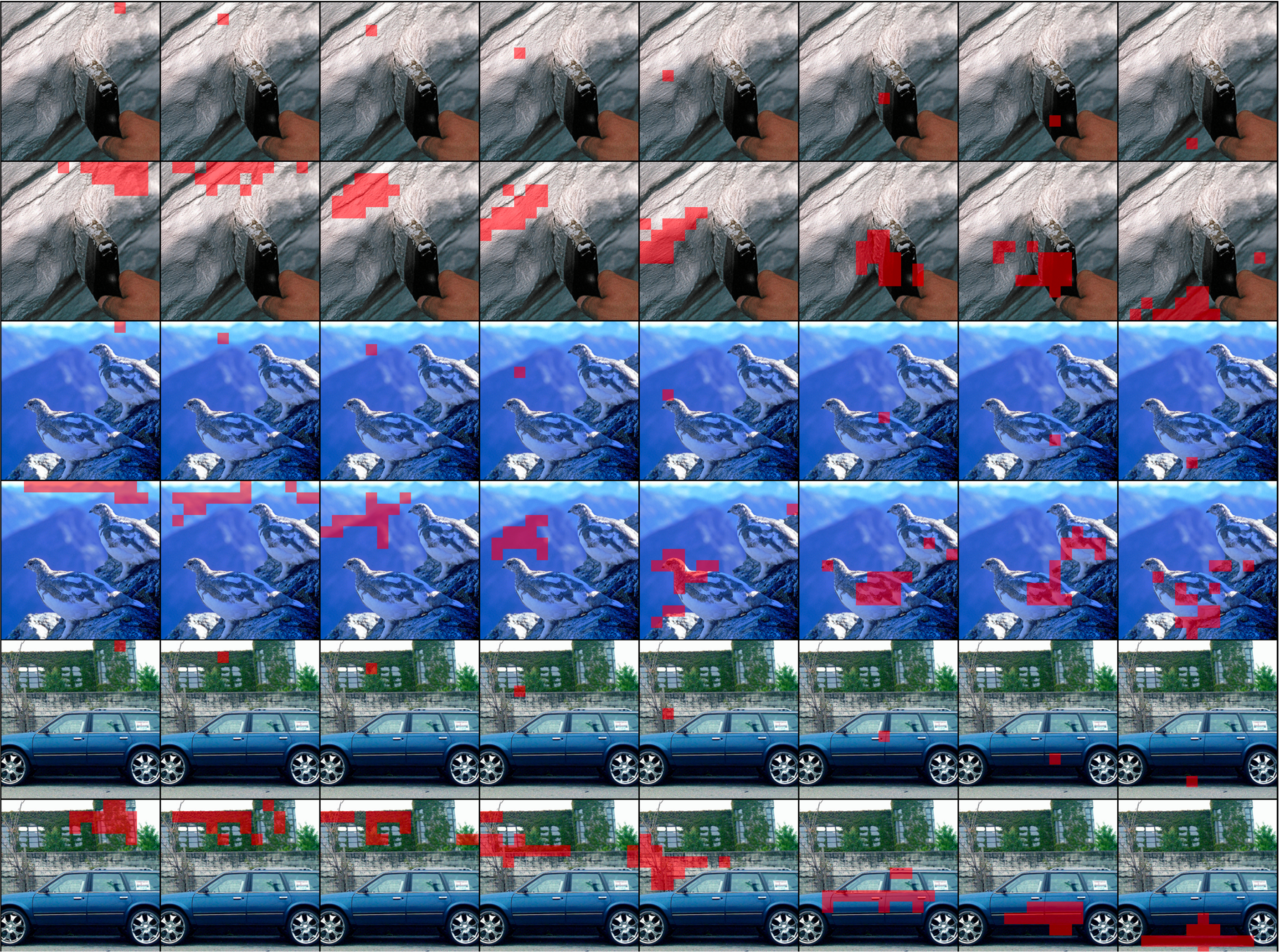}
\caption{Qualtitative visualization of learned cosine similarity maps given query patches using ViT-B/16. Rows for each sample denote the location of the given query patch and the top-10\% patches. Our DMT-JEPA performs effectively by encouraging the model to learn local semantics.
}
\label{fig: exp_vis_sim_three}
\end{figure*}

\begin{figure*}[t]
\centering
\includegraphics[width=0.9\linewidth]{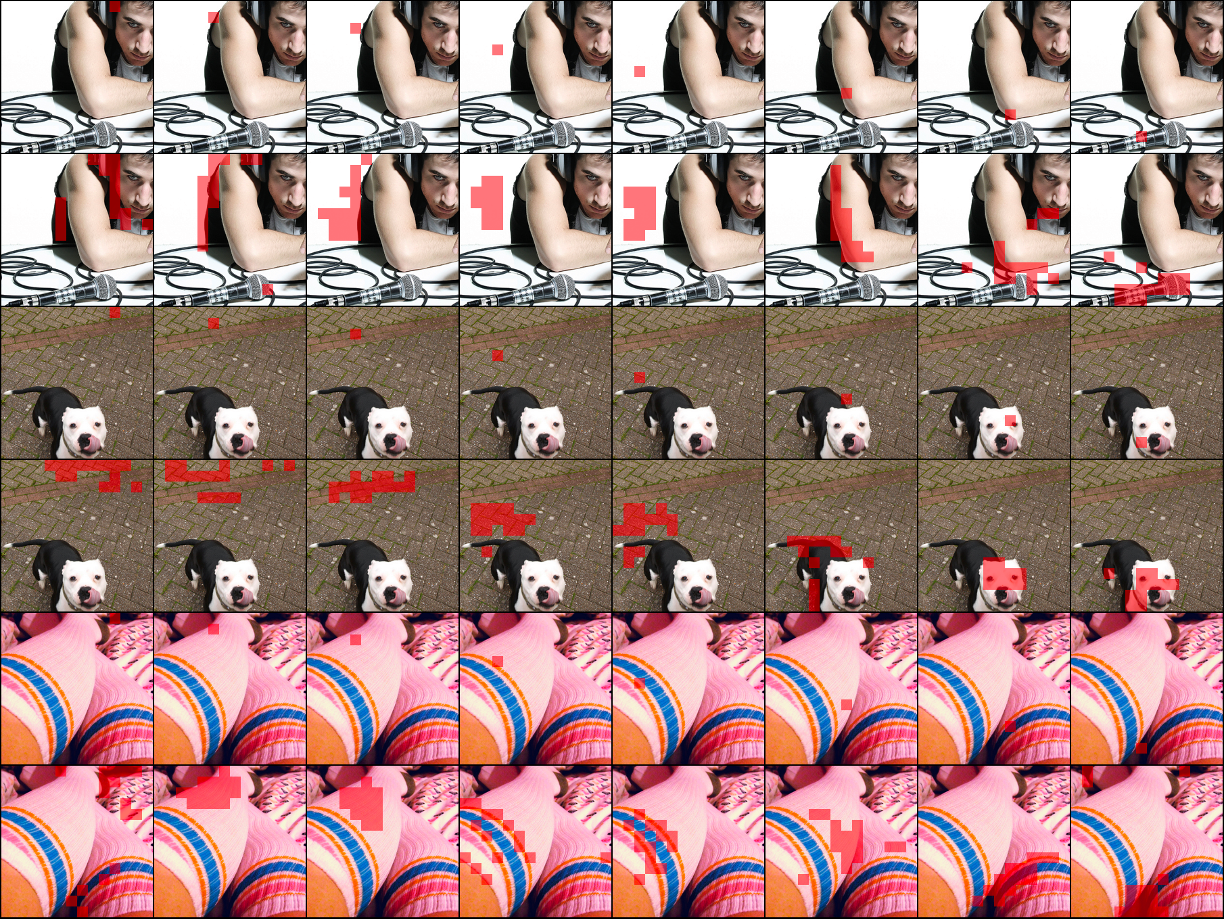}
\caption{Qualtitative visualization of learned cosine similarity maps given query patches using ViT-B/16. Rows for each sample denote the location of the given query patch and the top-10\% patches. Our DMT-JEPA performs effectively by encouraging the model to learn local semantics.
}
\label{fig: exp_vis_sim_four}
\end{figure*}

In order to qualitatively demonstrate the effectiveness of the proposed DMT-JEPA in learning local semantics during pre-training, we provide learned attention maps from the target encoder using pre-trained I-JEPA and our method, and attention maps from the cross-attention layer in the proposed Local Aggregation Target (LAT) modules in Figure~\ref{fig: exp_vis_attention1},~\ref{fig: exp_vis_attention2}, and~\ref{fig: exp_vis_attention3}.
Furthermore, we visualize the top-10\% of highly correlated patches in more examples by thresholding the cosine-similarity maps of query patches in the last layer in Figure~\ref{fig: exp_vis_sim_one},~\ref{fig: exp_vis_sim_two},~\ref{fig: exp_vis_sim_three}, and~\ref{fig: exp_vis_sim_four}.

\noindent\textbf{Learned Attention Maps.}
Learning discriminative attention maps is one of the key aspects of capturing local semantics for downstream tasks of dense prediction type, such as segmentation and detection.
To better evaluate the quality of learned attention maps, we visualize the averaged maps from different heads in the last attention layer by using a pre-trained ViT-B/16 target encoder.
For a more comprehensive comparison, we also add the attention maps from I-JEPA~\citep{assran2023self} target encoder and the cross-attention layer in our Local Aggregation Target (LAT) module.
The qualitative visualization results are shown in Figure~\ref{fig: exp_vis_attention1},~\ref{fig: exp_vis_attention2}, and~\ref{fig: exp_vis_attention3}.
Note that columns for each image sample represent the original image, attention maps from the target encoder in I-JEPA, attention maps from the target encoder in our DMT-JEPA, and attention maps from the local aggregation target module in our DMT-JEPA. 
As can be seen, both attention maps from the target encoder and the local aggregation target module in our DMT-JEPA are discriminative and focus on the object semantics in the image.
However, the attention maps from the target encoder in I-JEPA activate both the foreground and background and can not effectively discriminate the object semantics, because they did not incorporate local semantics into target representations as our DMT-JEPA did.
Meanwhile, the attention maps of the local aggregation target module in our DMT-JEPA have more focus on the details of foreground objects than that from the target encoder, indicating the effectiveness of our local aggregation target module in generating target representations with local semantics.
These high-quality visualization results further demonstrate the superiority of our new framework in learning meaningful representations with local semantics during self-supervised training, compared to I-JEPA, the state-of-the-art image-based joint-embedding predictive architecture.

\noindent\textbf{Learned Cosine Similarity Maps.}
To further validate the effectiveness of our method in learning discriminative dense representations, we visualize the top-10\% of highly correlated patches by thresholding the cosine similarity maps of query patches in the last layer using the pre-trained ViT-B/16 target encoder.
Figure~\ref{fig: exp_vis_sim_one},~\ref{fig: exp_vis_sim_two},~\ref{fig: exp_vis_sim_three}, and~\ref{fig: exp_vis_sim_four} showcase the qualitative visualization results, where rows for each sample denote the location of the given query patch and the top-10\% patches.
We can observe that the patches extracted by our DMT-JEPA are centralized and specifically focus on the location of the given query patch.
For example, given the first query patch on the building in the car example shown in Figure~\ref{fig: exp_vis_sim_three}, the top-10\% patches focus on the building.
When the query patch is given on the location of the car, the top-10\% patches also change to focus on the car.
Another example can also be seen in the dynamic changes with respect to the location of query patches on the head, arm, elbow, and microphone in the first sample shown in Figure~\ref{fig: exp_vis_sim_four}.
Interestingly, when the query patch is given on the body part from one of two birds in the second example shown in Figure~\ref{fig: exp_vis_sim_three}, the top-10\% patches can highlight the location of body parts from both birds, which might be due to the similar local semantics in both body locations.
These visualization results demonstrate that our DMT-JEPA performs effectively by encouraging the model to learn discriminative dense representations.

\end{document}